\documentclass[pmlr]{jmlr}% new name PMLR (Proceedings of Machine Learning)

\usepackage{multirow} 
\RequirePackage{graphicx}
 \usepackage{booktabs}
\usepackage{longtable}% for long tables
\makeatletter
\def\set@curr@file#1{\def\@curr@file{#1}} %temp workaround for 2019 latex release
\makeatother
\usepackage[load-configurations=version-1]{siunitx} % newer version

\theorembodyfont{\upshape}
\theoremheaderfont{\scshape}
\theorempostheader{:}
\theoremsep{\newline}

\jmlrvolume{252}
\jmlryear{2024}
\jmlrworkshop{Machine Learning for Healthcare}

\title[FineRadScore]{FineRadScore: A Radiology Report Line-by-Line Evaluation Technique Generating Corrections with Severity Scores}

\author{\Name{Alyssa Huang}
       \Email{alyssahuang@college.harvard.edu}\\ 
       \addr Harvard University \\
    \Name{Oishi Banerjee}
       \Email{oishi\_banerjee@g.harvard.edu}\\ 
       \addr Harvard University \\
    \Name{Kay Wu}
       \Email{kay.wu@medportal.ca}\\ 
       \addr Harvard University \\
    \Name{Eduardo Pontes Reis}
       \Email{eduardo.reis@einstein.br}\\ 
       \addr Stanford University \\
       \addr Hospital Israelita Albert Einstein \\
    \Name{Pranav Rajpurkar}
       \Email{pranav\_rajpurkar@hms.harvard.edu}\\ 
       \addr Harvard University \\
}

\begin{document}

\maketitle

\begin{abstract}
The current gold standard for evaluating generated chest x-ray (CXR) reports is through radiologist annotations. However, this process can be extremely time-consuming and costly, especially when evaluating large numbers of reports. In this work, we present FineRadScore, a Large Language Model (LLM)-based automated evaluation metric for generated CXR reports. Given a candidate report and a ground-truth report, FineRadScore gives the minimum number of line-by-line corrections required to go from the candidate to the ground-truth report. Additionally, FineRadScore provides an error severity rating with each correction and generates comments explaining why the correction was needed. We demonstrate that FineRadScore's corrections and error severity scores align with radiologist opinions. We also show that, when used to judge the quality of the report as a whole, FineRadScore aligns with radiologists as well as current state-of-the-art automated CXR evaluation metrics. Finally, we analyze FineRadScore's shortcomings to provide suggestions for future improvements.
\end{abstract}

\section{Introduction}

Artificial Intelligence (AI) is rapidly advancing in the field of medical image interpretation. In particular, models have succeeded on a range of chest X-ray (CXR) classification tasks, improving radiologists' efficiency (\cite{medical-ai-efficiency}). Models performing the more comprehensive task of radiology report generation could further assist radiologists by describing all abnormalities in a chest X-ray in fluent text (\cite{cxr-repair}, \cite{x-rem}, \cite{warm-starting}). However, AI models still struggle to produce reliable and accurate radiology reports. As a result, there is a need for evaluation metrics that can track progress in this space and provide detailed feedback on AI-generated reports.

Automating the evaluation of AI-generated reports is a challenging task. The current gold standard is manual evaluation by radiologists, yet evaluating a large number of reports in this manner is costly and time-consuming, creating a need for automated metrics. Researchers have explored natural language generation (NLG) metrics such as BLEU (\cite{bleu}) and BERTScore (\cite{bert-score}) to compare the generated report with a ground-truth report. However, these metrics were designed for non-medical text and fail to capture important features of medical text due to the lack of domain-specific knowledge needed to evaluate radiology reports (\cite{evaluating-progress-patterns}, \cite{gpt4-better-human-alignment}, \cite{adams2023metaevaluationfaithfulnessmetricslongform}). For example, if a model generated a sentence such as ``there is a focal lesion'' and the ground-truth report stated ``there is no focal lesion,'' this pair would score highly on BLEU-1, even though radiologists would consider it a poor match.

As a result, recent works have developed metrics specifically designed for evaluating radiology reports, such as CheXbert vector similarity (\cite{chexbert-vector-similarity}), RadGraph-F1 (\cite{radgraph-f1}), and RadCliQ (\cite{evaluating-progress-patterns}). Although these metrics have shown promise, they focus on report-level evaluation, outputting a single score for the entire report. This work expands on these existing automated evaluation metrics by using large language models to offer more detailed assessments, with sentence-level feedback and fine-grained estimates of error severity. 

\begin{figure}
\includegraphics[width=\textwidth]{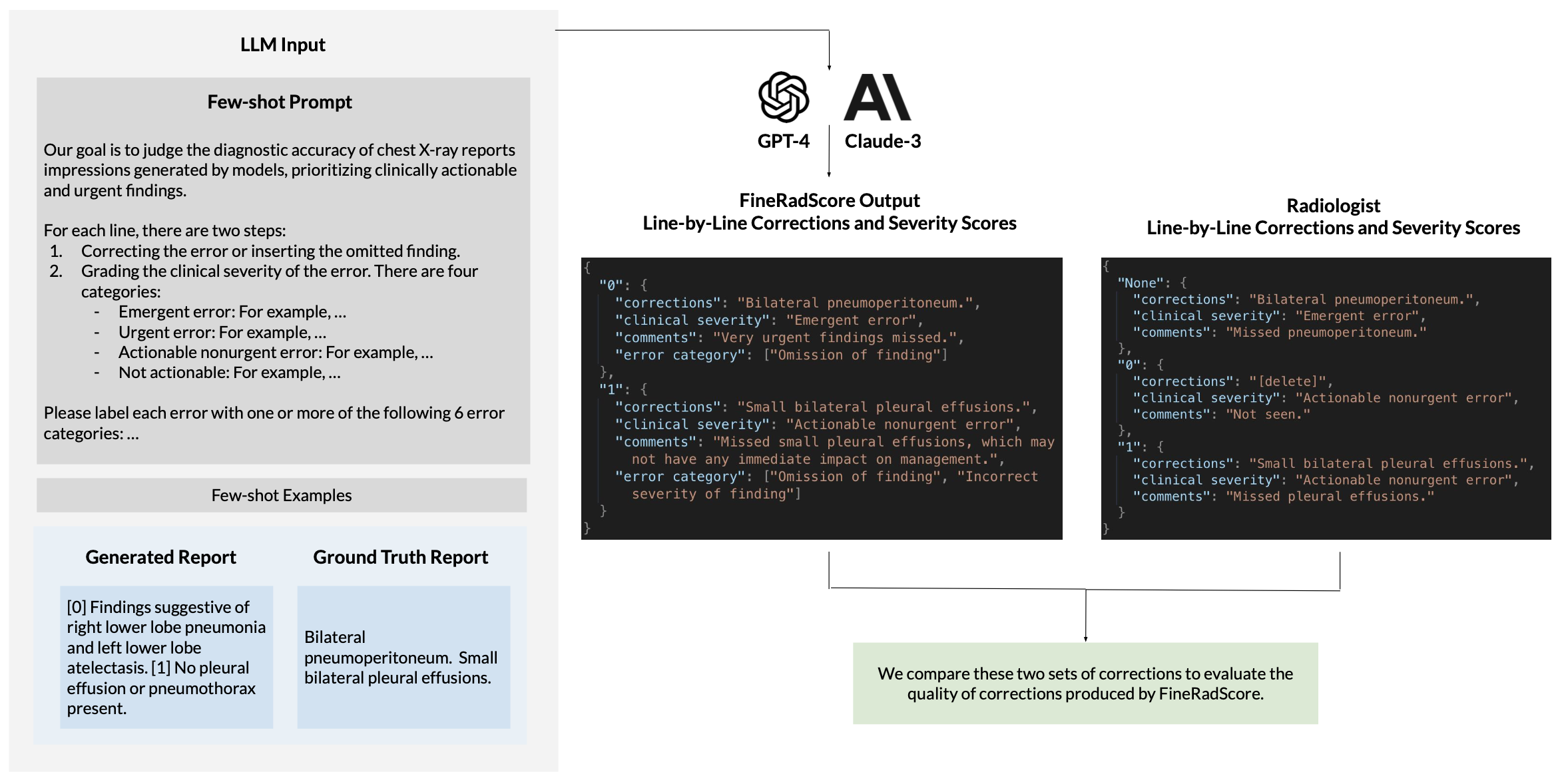}
\caption[Overall Evaluation Framework for FineRadScore.]{Overall Evaluation Framework for FineRadScore. We first prompt an LLM with a candidate report and a ground-truth report and ask for line-by-line corrections, corresponding clinical severity ratings, and comments describing each error that is corrected. Then, we feed in the pair of reports: one candidate and one ground truth. We then extract the FineRadScore-generated corrections, severity scores, and comments, and compare them with radiologist annotations. In this paper, we examine two different LLMs as backbones for the FineRadScore framework: GPT-4 and Claude-3 Opus. We will refer to FineRadScore-GPT-4  and FineRadScore-Opus to distinguish between both versions.

\label{fig:fineradscore-framework}}
\end{figure}

\subsection*{Generalizable Insights about Machine Learning in the Context of Healthcare}
In this work, we propose an evaluation framework that uses a Large Language Model (LLM) to offer sentence-level corrections, error severity ratings, and comments. We compare and contrast performance between using GPT-4 and Claude-3 Opus as our framework's backbone. To our best knowledge, this is the first work to:
\begin{enumerate}
    \item Develop a CXR evaluation framework that can also generate corrections
    \item Develop a CXR evaluation framework that assigns clinical severity scores and error types found for each line
    \item Apply and analyze Claude-3 Opus on any radiology report evaluation task.
\end{enumerate}  

To help evaluate our proposed framework, we also present a dataset of 100 studies with detailed sentence-level radiologist annotations. Using this new data and other existing datasets, we show that our proposed evaluation technique aligns with radiologists' judgments, as well as identify where there is still room for improvement. 

\textbf{Motivation for Line-by-Line Evaluation.} Combined with information on clinical error severity, the number of incorrect lines in a report helps provide a well-rounded picture of how an AI model would impact clinical practice. In particular, clinical error severity correlates with consequences on patient health, while the number of incorrect lines correlates with time needed by radiologists to correct the report. Line-by-line corrections can provide insight into which parts of a report are most flawed (e.g. ``the impressions section is consistently poor-quality'' or ``lines mentioning fractures are frequently incorrect''). Additionally, line-by-line corrections could be used to improve models through Reinforcement Learning with AI Feedback techniques.

While it can also be used as an intermediate step to generate preference data for RLHF frameworks, FineRadScore needs both the candidate and ground truth reports as input. Therefore, it should still primarily be used as an evaluation metric.

\section{Related Work} 

\subsection{Automated Evaluation Techniques for AI-Generated Radiology Reports.}

NLG metrics such as BLEU (\cite{bleu}) and BERTScore (\cite{bert-score}) are popular choices when evaluating AI-generated chest X-ray reports (\cite{evaluating-progress-patterns}, \cite{x-rem}). BLEU measures token overlap between reports, while BERTScore compares embeddings generated by a BERT-based model that is not specific to the medical domain. In addition, other metrics specific to the medical field such as CheXbert vector similarity (\cite{chexbert-vector-similarity}), RadGraph-F1 (\cite{radgraph-f1}), and RadCliQ (\cite{evaluating-progress-patterns}) have been developed and been widely used to evaluate chest x-ray reports (\cite{towards-generalist-biomedical-ai}, \cite{maira-1}). CheXbert vector similarity takes in a report and produces a 14-dimensional vector, where each index indicates whether a certain pathology was found in the report, and compares cosine similarities between vectors. RadGraph-F1 extracts clinical entities and relations to create a knowledge graph from a report and computes the overlap with a ground-truth graph. RadCliQ is a weighted composite of all the aforementioned metrics (BLEU, BertScore, CheXbert, and RadGraph-F1). Among these metrics, RadCliQ has been found to be most aligned with radiologist opinions (\cite{evaluating-progress-patterns}).

Although these metrics can score the overall quality of a report using clinical knowledge, they are unable to give more fine-grained information. For example, they fail to explain which parts of a report are problematic or how clinically significant an error is. Additionally, they do not indicate how the generated report can be edited to make its content align with the ground-truth report's. This work aims to fill these gaps.

\subsection{GPT-4 as an Evaluator.} Recent works have found LLMs, GPT-4 in particular, to be promising in evaluating NLG outputs. G-Eval, a GPT-4-based metric, evaluates NLG outputs using chain-of-thought prompting (\cite{gpt4-better-human-alignment}). For a pair of candidate and ground-truth reports, G-Eval assigns a score based on the quality of the candidate report. The authors found that G-Eval is more aligned with human judgement than BLEU and BERTScore. In the medical domain, G-Rad is a similar technique, which prompts GPT-4 to identify the number of clinically significant and insignificant errors in a given report according to a set of error categories (\cite{chaves2024training}). Additionally, entailment-based prompts have been used on LLMs in the medical space to obtain a single score for each pair of reports (\cite{gpt4-fine-grained-eval}, \cite{gpt4-medical-texts}). 

We expand on these methods in our work by exploring whether GPT-4 can go beyond just identifying the number of errors to generating the corrections for those errors. Additionally, our work maintains the ability to detect the significance of each error by also producing a clinical severity rating for each error. Finally, we explore one of the Claude-3 models, which to the best of our knowledge, has not yet been examined in this task.

\subsection{Alignment Datasets.} In order to measure how well automated metrics align with human assessments of report quality, past work has curated datasets containing radiologist scores of chest X-ray reports. Evaluation scores from automatic metrics can be compared with radiologist scores on these datasets. We introduce two such datasets, which we will use to measure how well FineRadScore aligns with human evaluations.

\textbf{ReFiSco-v0.} ReFiSco-v0 (\cite{refisco-v0}) includes 60 chest X-ray studies, each containing several candidate  reports with line-by line expert annotations. For each line in a candidate report, radiologists were asked to either state that the line was already correct or else make a correction by rewriting or deleting the line. They were also allowed to make corrections to a report by inserting new lines. In addition, they were asked to rate each error they corrected using five clinical severity categories: `No error'', ``Not actionable'', ``Actionable nonurgent error'', ``Urgent error'', or ``Emergent error'' in order of increasing severity.

\textbf{ReXVal.} ReXVal (\cite{evaluating-progress-patterns}) is an expert-annotated set of 200 studies, each including an AI-generated candidate report and a radiologist-written ground-truth report. Radiologists identified errors in each candidate report by comparing it to the ground-truth report. They classified errors into six error categories: 1) False prediction of finding, 2) Omission of finding, 3) Incorrect location/position of finding, 4) Incorrect severity of finding, 5), Mention of comparison that is not present in the reference impression, 6) Omission of comparison describing a change from a previous study. When evaluating each candidate report, radiologists were asked to count the number of clinically significant and insignificant errors in each of the six error categories. In our work, when we reference these six error categories, we will refer to them as the ``ReXVal error categories'' to explicitly contrast them from the clinical severity categories introduced in ReFiSco-v0.

\section{Methods}

\subsection{FineRadScore Evaluation Framework} \label{sec:gpt4-evaluation-framework}

In Figure \ref{fig:fineradscore-framework}, we see an overview of FineRadScore. We feed in a pair of reports: the candidate report and a ground-truth report. Our goal is to have the LLM correctly identify the line-by-line corrections needed to make the content of the candidate report match the ground-truth report's using the least number of changes possible. For each correction, the LLM should mark the correction as involving either the deletion, substitution, or insertion of a line. Additionally, the LLM should label the correction with one of five clinical error severity categories: ``Not actionable'', ``Actionable nonurgent error'', ``Urgent Error'', ``Emergent Error'', ``Invalid Comparison'', and provide comments regarding why the correction is needed. We note that these clinical error severity categories are identical to ReFiSco-v0's with the addition of ``Invalid Comparison,'' whose motivation we discuss below. To help users gain a better understanding of the types of errors being produced, we also ask it to output a list of error categories for each error it finds. We use the ReXVal error categories: ``False prediction of finding'', ``Omission of finding'', ``Incorrect location/position of finding'', ``Incorrect severity of finding'', ``Mention of comparison that is not present in the reference impression'', ``Omission of comparison describing a change from a previous study.'' The list produced by FineRadScore may contain 0-6 error categories, as multiple error types may appear in a given line. In the body of this work, we primarily analyze how the corrections and clinical severity scores outputted by FineRadScore align with radiologists. Analyses of comments and the list of error labels according to the ReXVal categories are left to the appendix.

\subsection{LLM Prompt}
Our prompt describes the task, instructing the model to only include semantically relevant errors and to ignore stylistic differences. To construct a few-shot prompt, we then include five examples from our radiologist annotations, shown in Appendix \ref{appendix:few-shots}. We also tried using a zero-shot prompt shown in \ref{appendix:zero-shot}, but found that it consistently underperformed the few-shot prompt. Thus, we omit zero-shot results in this paper. The full prompt is shown in Appendix \ref{appendix:prompt-instructions}.

\subsection{Intended JSON Output}
We ask the LLM to output a list of corrections in JSON format shown in Appendix \ref{appendix:few-shots}. In practice, not every generation adhered to this format, so we would re-prompt the LLM up to a certain number of retries if the output was not formatted properly.

\section{ReFiSco-v1 Dataset}

The goal of the ReFiSco-v1 dataset is to ensure that FineRadScore can correct and score reports in a way that aligns with radiologists. The dataset designed is based off of the ReFiSco-v0 dataset. In this section, we describe how we created the ReFiSco-v1 dataset. 

\subsection{Cohort Selection}
To start, we sample 100 radiology studies from the MIMIC-CXR test dataset (\cite{mimic-cxr}), spanning a total of 14 subjects.  We next use ClsGen (\cite{clsgen}), a model with high clinical accuracy on MIMIC-CXR that is available to researchers, to produce our candidate reports. 

For each study in our sample, we produce two reports: a findings report and an impressions report. First, we use ClsGen to generate the report findings. We then use GPT-4 to produce the impressions as a summarization task. We treat the findings reports and the impressions reports separately.

\subsection{Annotation Process}
After obtaining these 2 reports for each study, we recruited 2 radiologists to annotate subsets of the selected radiology studies. One of the radiologists is board-certified with 7 years of experience, and the other is a resident. Both radiologists were given a 30-minute orientation, as well as instructions and examples to complete their annotation tasks.

The radiologists were presented with all chest X-ray images from a given study in the MIMIC-CXR dataset. Then, they were asked to annotate each report line by line. Each annotation consisted of two parts. First, the radiologist was asked to correct the line by either deleting an existing line, substituting an existing line for a new line, or inserting a new line. Then, they are asked to mark each line of the report with a clinical severity rating. In order of increasing severity, the categories are ``No error'', ``Invalid comparison'', ``Not actionable'', ``Actionable nonurgent error'', ``Urgent error'', or ``Emergent error''. Descriptions and examples for each of these errors were given in the instructions handout. These severity categories are the same as ReFiSco-v0's, with the exception of ``Invalid comparison,'' a new category that ReFiSco-v0 grouped together with ``Not actionable''. Since these models are trained on real-world datasets that make references to priors, they often make hallucinated references to priors that should be removed (\cite{cxr-remove-priors}). The ``Invalid comparison'' error category captures this failure mode. On lines where both a hallucinated prior and another error are found, we ask radiologists to rate the severity of the other error.

\subsection{ReFiSco-v1 vs ReFiSco-v0}

ReFiSco-v1 is similar to ReFiSco-v0 and collects radiologist annotations using the same set of instructions. The key distinctions between ReFiSco-v0 and ReFisco-v1 are that: 1) ReFiSco-v1 considers generations from the AI model ClsGen (\cite{clsgen}) while ReFiSco-v0 considers generations from AI models X-REM (\cite{x-rem}) and CXR-RePaiR (\cite{cxr-repair}) and 2) ReFiSco-v1 considers both findings and impression generations while ReFiSco-v0 only considers impressions.

\section{Experiments}
\label{experiments}

To assess the efficacy of FineRadScore, we treat the radiologist-corrected report version of the candidate report as the ground truth. To do this, we apply the corrections the radiologists made to the model-generated report to produce our ground-truth report. Then, we feed each pair of reports, consisting of a model-generated candidate report and its radiologist-corrected counterpart, as input into our evaluation pipeline. There were some AI-generated reports that were annotated by multiple raters. We create different pairs with each of these reports and submit each radiologist-corrected version to the LLM separately. 

For each experiment, we use the evaluation framework outlined in section \ref{sec:gpt4-evaluation-framework}. Sometimes, the LLM is not able to output the correct JSON format we asked for. In those cases, we regenerate the result with up to 5 retries. In almost all of the cases, this number was sufficient to obtain a correctly-formatted output. 

We decide to evaluate FineRadScore on both ReFiSco-v0 and ReFiSco-v1. ReFiSco-v0 and ReFiSco-v1 collectively contain candidate reports from four different sources (ClsGen (\cite{clsgen}), X-REM (\cite{x-rem}), CXR-RePaiR (\cite{cxr-repair}), and the MIMIC expert report (\cite{mimic-cxr})). The two datasets also cover two common report sections: findings and impressions. In order to ensure that FineRadScore generalizes well across models and report types, we choose to evaluate on both of these datasets. Through our experiments, we explore whether:
\\
\indent 1. FineRadScore correctly identifies the type of corrections (e.g. deletion, rewriting) needed for each line.
\\
\indent 2. FineRadScore is able to correctly generate the new text that should be substituted in or inserted.
\\
\indent 3. FineRadScore is able to determine the clinical severity rating of each error.
\\
\indent 4. When we apply FineRadScore's corrections to our candidate report, we end up with a report that is more aligned with the ground truth.
\\
\indent 5. Results 1-4 hold when the candidate report is stylistically different (shuffled or paraphrased) from the ground-truth report.
\\
\indent 6. FineRadScore performs as well as RadCliQ (\cite{evaluating-progress-patterns}) and G-Rad (\cite{chaves2024training}), two current state-of-the-art techniques, when scoring a report's overall quality. 

\section{Results}

All results shown below are generated using the few-shot prompt.

\subsection{FineRadScore Correction Type Quality}
\label{sec:correction-type}

We first study whether FineRadScore is able to identify the correction type needed for each line. To do so, we define categories describing the type of correction for each line in the AI-generated report: 1) No Change, 2) Delete, 3) Rewrite, or 4) Insert. The first three categories affect the original line in the generated report, while the fourth adds a new line that did not originally exist. Figure \ref{fig:5.1_correction_type} in the appendix illustrates how often the radiologist performs each type of correction in ReFiSco-v1 and ReFiSco-v0. 

Within each category, we measure how often FineRadScore makes the correct type of correction, matching the radiologist. For example, for each line that a radiologist chooses not to change, we measure how often FineRadScore also makes no change. Likewise, for each line that a radiologist chooses to delete or rewrite, we measure how often FineRadScore makes the same decision on that line. 

We treat insertions as a special case, as the same content can be inserted using different numbers of lines. For instance, a radiologist may insert three lines (``No focal consolidation. No pleural effusion. No pneumothorax.''), while FineRadScore may insert the exact same content using one line (``There is no focal consolidation, pleural effusion or pneumothorax''). We therefore analyze insertions at the level of whole reports, not individual lines. In other words, whenever a radiologist inserts at least one new line into a report, we measure how often FineRadScore also inserts at least one line into that report. We also measure FineRadScore's precision while inserting lines, by examining how often FineRadScore does not insert a line when the radiologist does not insert any.

In this experiment, we feed in the following pair of reports at each time step. \textbf{Candidate Report}: Original Generated Report. \textbf{Ground-Truth Report}: Radiologist-Corrected Generated Report

\begin{table}[htbp]
    \centering
    \caption[FineRadScore Accuracy by Correction Type in Original Setting.]{FineRadScore is able to identify the type of correction the radiologist makes at each line in the report or in the report as a whole. }
    \label{tab:correction-type-accuracies}
    \begin{tabular}{ccccccc}
    \toprule
    & & \multicolumn{3}{c}{Line Level Correction} & \multicolumn{2}{c}{Report Level Correction} \\
    \midrule
    \bfseries Dataset & \bfseries Model & \bfseries No Change & \bfseries Delete & \bfseries Rewrite & \bfseries Insert & \bfseries No Insert \\
    \midrule
    \multirow{ 2}{*}{ReFiSco-v1} & GPT-4 & \bfseries 94.85\% & 80.60\% & 86.80\% & 68.75\% & 100\% \\
    & Opus & 90.86\% & \bfseries 97.39\% & \bfseries 94.00\% & \bfseries 93.75\% & 100\% \\ 
    \midrule
    \multirow{ 2}{*}{ReFiSco-v0} & GPT-4 & \bfseries 91.18\% & 68.85\% & 87.07\% & 72.15\% & 100\% \\
    & Opus & 73.38\% &\bfseries  88.52\% & \bfseries 91.81\% & \bfseries 94.94\% & 100\% \\ 
    \bottomrule
    \end{tabular}
\end{table}

As shown in Table \ref{tab:correction-type-accuracies}, FineRadScore-GPT-4 can reliably identify when a line should be left unchanged, but its performance drops when lines need to be deleted or inserted. Interestingly, in reports where a radiologist deletes a line and then makes an insertion, FineRadScore-GPT-4 often chooses to instead rewrite the line in one step, leading to low performance in identifying insertions. As a result, FineRadScore-GPT-4 may still produce a final corrected text that matches the ground truth, even though it makes different types of corrections. Overall, FineRadScore-Opus performs most consistently, achieving accuracy rates of over 90\% on all correction types on ReFiSco-v1. No model inserted new lines when the radiologist did not insert any lines across both datasets, indicating robustness to this failure mode.

\subsection{FineRadScore Text Rewrite and Insertion Quality}
\label{sec:text-similarity}

Next, we compare the lexical similarity between the new text proposed by radiologists and FineRadScore when rewriting or inserting lines. We aim to ensure that when FineRadScore decides to rewrite or insert a line, it does so in a way matching the radiologist. In Table \ref{tab:gpt4-rewrite-insertion-text}, we compare the concatenation of all rewritten and inserted lines done by FineRadScore and the concatenation of all rewritten and inserted done by the radiologist for each report. We choose to compare the concatenations because FineRadScore may choose to correct a report using different steps as the radiologist but still arrive at the same final result. For example, for a sample report ABC (original) $\rightarrow$ ABD (corrected), FineRadScore may choose to first delete C, then insert D, while the radiologist just rewrites C as D. As shown in Table \ref{tab:gpt4-rewrite-insertion-text}, FineRadScore is able to capture a majority of the text either rewritten or inserted by radiologists in its corrections.

\begin{table}[htbp]
    \centering
    \caption[FineRadScore Rewrite and Insert BLEU scores in Original Setting.]{FineRadScore is able to capture a majority of the text either rewritten or inserted by radiologists in its corrections.}
    \label{tab:gpt4-rewrite-insertion-text}
    \begin{tabular}{ccccc}
    \toprule
    \bfseries Dataset & \bfseries Model &  \bfseries BLEU-1 Scores & \bfseries BLEU-2 Scores & \bfseries BERT Scores \\
    \midrule
    \multirow{ 2}{*}{ReFiSco-v1} & GPT-4 & \bfseries 0.8601 & \bfseries 0.8191 & \bfseries 0.9503 \\
    & Opus & 0.8590 & 0.8101 & 0.9418 \\ 
    \midrule
    \multirow{ 2}{*}{ReFiSco-v0} & GPT-4 & 0.7776 & 0.7398 & 0.8980 \\
    & Opus & \bfseries 0.8371 & \bfseries 0.8095 & \bfseries 0.9175 \\ 
    \bottomrule
    \end{tabular}
\end{table}

\subsection{Correcting Candidate Reports Using FineRadScore}
\label{sec:original-correcting-results}

We now aim to ensure that after applying FineRadScore's corrections to our candidate report, we obtain a corrected report that is more aligned with our ground-truth report. In particular, we wanted to see if a FineRadScore-corrected report exhibited 1) more text similarity and 2) more semantic similarity to the ground truth reports. Therefore, we went through each correction outputted by FineRadScore and applied it to the original line in the candidate report to obtain a corrected report. We then compare the FineRadScore-corrected report and the ground-truth report in three ways. First, we compare the corrected candidate report and the ground-truth report's text similarities using BLEU-2 scores. However, as noted in the introduction, lexical similarity need not be correlated with semantic alignment. Therefore, we use RadGraph-F1 (\cite{radgraph-f1}) and RadCliQ (\cite{evaluating-progress-patterns}), two evaluation metrics designed for chest x-ray reports and which have been shown to align well with radiologists. RadGraph-F1 measures the similarity between the sets of medically relevant words and their relationships for a pair of reports and outputs a score. The score ranges from 0 to 1, with 1 representing the highest alignment between the reports. RadCliQ is a composite metric that combines both lexical and clinical metrics for a pair of reports to output a score. The lower the score, the more aligned the pair of reports are.

To establish a baseline, we start by comparing the original uncorrected model-generated reports with the ground-truth reports using the three methods described above. Then, we compare the FineRadScore-corrected reports with the ground-truth reports to see if our corrected candidate reports are more aligned with the ground truth than the uncorrected candidate reports.

As shown in Table \ref{tab:original-radcliq}, when we apply the corrections generated by FineRadScore to our candidate report, we obtain higher BLEU scores when comparing this corrected report to our ground-truth report. This finding is consistent across models and datasets and is most prominent with FineRadScore-Opus. FineRadScore also produces corrected candidate reports with higher RadGraph-F1 scores and lower RadCliQ scores. Therefore, according to these metrics, FineRadScore is producing higher-quality reports that are more medically aligned with the ground-truth reports.

\begin{table}[htbp]
    \centering
    \caption{FineRadScore produces corrected candidate reports with higher RadGraph-F1 scores and lower RadCliQ scores. Therefore, FineRadScore is producing higher-quality reports that more aligned with the ground-truth reports.}
    \label{tab:original-radcliq}
    \begin{tabular}{cccccccc}
    \toprule
    & & \multicolumn{2}{c}{BLEU-2 Scores} & \multicolumn{2}{c}{RadGraph-F1} & \multicolumn{2}{c}{RadCliQ} \\
    \midrule
    \bfseries Dataset & \bfseries Model & \bfseries Base & \bfseries Corrected & \bfseries Base & \bfseries Corrected & \bfseries Base & \bfseries Corrected \\
    \midrule
    \multirow{ 2}{*}{ReFiSco-v1} & GPT-4 & \multirow{ 2}{*}{0.4232} & 0.9005 & \multirow{ 2}{*}{0.5376} & \bfseries 0.9395 & \multirow{ 2}{*}{0.0118} & -1.245 \\
    & Opus & & \bfseries 0.9134 & & 0.9391 & & \bfseries -1.249 \\ 
    \midrule
    \multirow{ 2}{*}{ReFiSco-v0} & GPT-4 & \multirow{ 2}{*}{0.3994} & 0.8618 & \multirow{ 2}{*}{0.4633} & 0.8735 & \multirow{ 2}{*}{0.2285} & -1.143 \\
    & Opus & & \bfseries 0.9004 & & \bfseries 0.8935 & & \bfseries -1.192 \\ 
    \bottomrule
    \end{tabular}
\end{table}

\subsection{FineRadScore Clinical Severity Ratings}
\label{sec:clinical-severity-ratings}

We next study whether FineRadScore is able to estimate the clinical severity of each error it corrects. Following \cite{refisco-v0}, we treat unwanted comparisons to priors as not actionable errors in quantitative analyses. When no corrections are found, we set the clinical severity rating to 0. Then, for each of the four severity ratings, we assign the numbers 1 to 4 in order of increasing severity (not actionable, actionable nonurgent error, urgent error, emergent error). We note that there are multiple valid ways to get a report-level score from line-level scores and that our paper explores both taking the maximum line-level score and summing all line-level scores. In this section, we take the maximum clinical severity over all lines in a given report to produce a clinical severity score. We then compare FineRadScore's report-level clinical severity score with the radiologist's.

To provide a baseline for how well FineRadScore captures the severity rating for each line, we also measure how often different radiologists agree with each other, using inter-rater disagreement to provide a baseline. Inter-rater disagreement gives us a sense of the best we can expect from FineRadScore. For example, if all raters find the same error but are split between whether the error is emergent or urgent, the best we can expect from FineRadScore is to find the same error and rate it within that range of severities. In ReFiSco-v1, we did not collect annotations from different radiologists for a single report, so we cannot compute inter-rater disagreement. However, in ReFiSco-v0, for generations with multiple radiologist ratings, we randomly choose one of the radiologist’s ratings to be the candidate and treat all other radiologists’ ratings as ground-truth ratings. We then compare the candidate radiologist's severity ratings to the ground truth radiologists' to obtain inter-rater disagreement scores on ReFiSco-v0.

\begin{figure}[!htb]
   \begin{minipage}{0.48\textwidth}
     \centering
     \includegraphics[width=\linewidth]{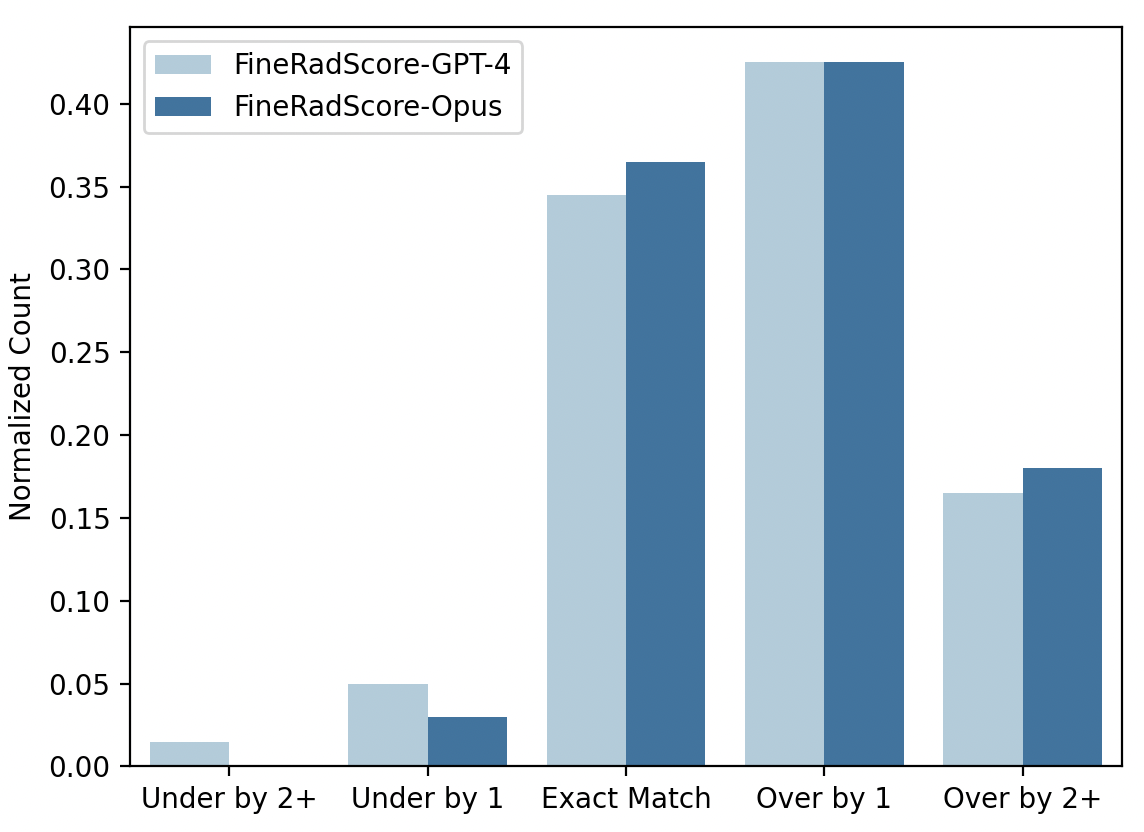}
   \end{minipage}\hfill
   \begin{minipage}{0.48\textwidth}
     \centering
     \includegraphics[width=\linewidth]{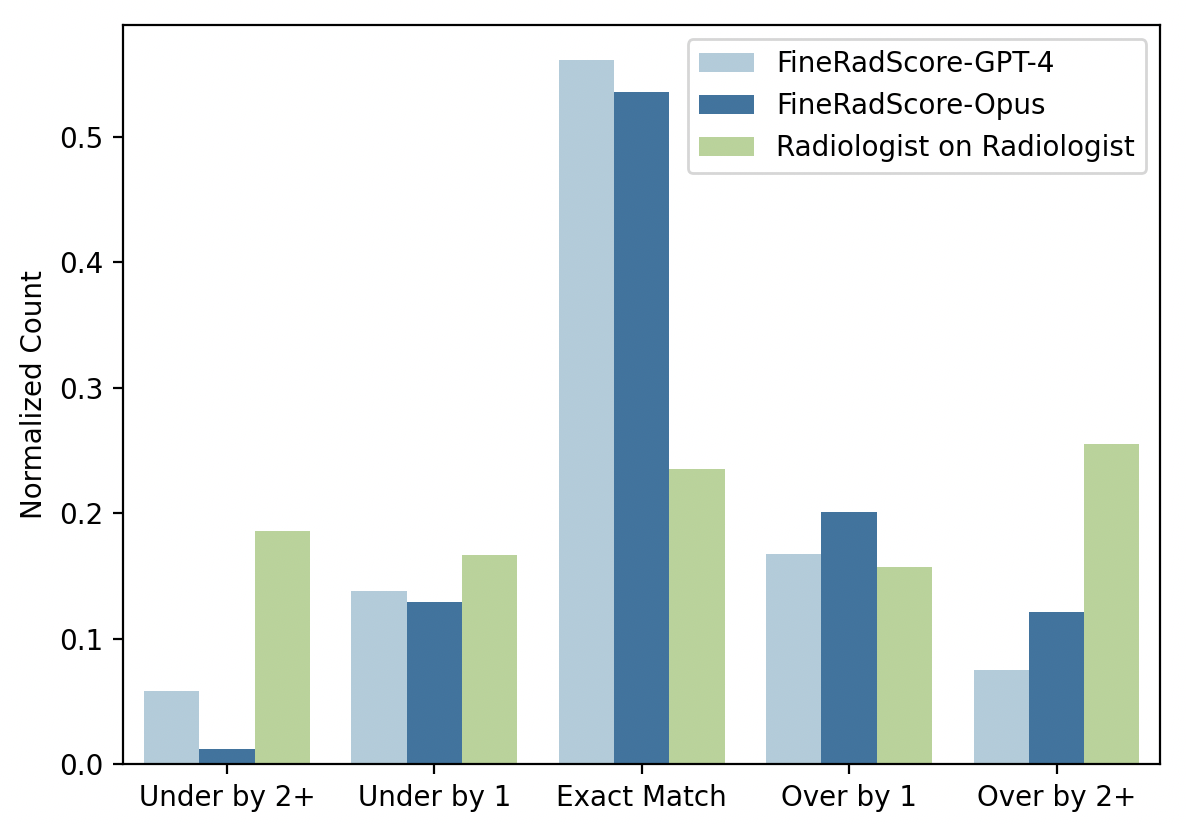}
   \end{minipage}
   \caption{FineRadScore Clinical Severity Scores' Accuracies on ReFiSco-v1 (left) and ReFiSco-v0 (right). FineRadScore's is able to determine clinical severity scores within an error range of one category.}
   \label{fig:original-model-comp}
\end{figure}

In Figure \ref{fig:original-model-comp}, we plot the errors between radiologists' severity ratings and FineRadScore severity ratings below for both ReFiSco-v1 and ReFiSco-v0. We also plot the inter-rater disagreement for ReFiSco-v0. Because the number of reports are different between FineRadScore and the radiologists, we divide by the number of reports.

In both graphs, there is a heavy concentration in the center for the errors found by FineRadScore, which indicates that in most cases, FineRadScore has the same severity rating as the ground truth radiologist or was off by one. FineRadScore has stronger performance in rating clinical error severities in ReFiSco-v0 compared to ReFiSco-v1. We also note that FineRadScore appears to have better alignment with the ground truth radiologist compared to a candidate radiologist. However, this comparison may not be fair to the candidate radiologist since FineRadScore has access to the report corrected by the ground truth radiologist while the candidate radiologist does not. Nevertheless, these graphs indicate that FineRadScore is still able to identify the severity of each error to a significant degree. The mean absolute errors are all under 1 for FineRadScore, at around 0.84 for ReFiSco-v1 and 0.62 for ReFiSco-v0.

\section{Stylistically Dissimilar Generated and Ground-truth reports}

Up to this point, we have been working with candidate and ground-truth reports that are stylistically similar to each other. When moving from the candidate report to the ground-truth report, we asked radiologists to make the minimum number of corrections. Therefore, our candidate reports are stylistically similar to our ground-truth reports. However, this assumption does always hold in practice. Therefore, we now test whether the results above hold when the candidate and ground-truth reports are stylistically dissimilar.

\subsection{A Shuffled Ground Truth}

We start by shuffling the order in which findings are reported in the ground truth. Then, we feed in the original candidate report with the permuted ground-truth report instead of the original ground-truth report and replicate the experiments above. 

\begin{table}[htbp]
    \centering
    \caption[FineRadScore Accuracy by Correction Type in Shuffled Setting.]{FineRadScore is able to identify the type of correction the radiologist makes at each line in the report or in the report as a whole even when the ground-truth report is shuffled. }
    \label{tab:shuffled-correction-type}
    \begin{tabular}{ccccccc}
    \toprule
    & & \multicolumn{3}{c}{Line Level Correction} & \multicolumn{2}{c}{Report Level Correction} \\
    \midrule
    \bfseries Dataset & \bfseries Model & \bfseries No Change & \bfseries Delete & \bfseries Rewrite & \bfseries Insert & \bfseries No Insert \\
    \midrule
    \multirow{ 2}{*}{ReFiSco-v1} & GPT-4 & \bfseries 92.90\% & 82.46\% & 91.60\% & 80.21\% & 100\% \\
    & Opus & 80.35\% & \bfseries 95.15\% & \bfseries 92.80\% & \bfseries 95.83\% & 100\% \\ 
    \midrule
    \multirow{ 2}{*}{ReFiSco-v0} & GPT-4 & \bfseries 86.18\% & 68.85\% & 86.64\% & 62.03\% & 100\% \\
    & Opus & 61.62\% & \bfseries 89.34\% & \bfseries 92.24\% & \bfseries 88.61\% & 100\% \\ 
    \bottomrule
    \end{tabular}
\end{table}

We first examine whether FineRadScore is still able to identify the type of correction needed at each line in this shuffled setting. In Table \ref{tab:shuffled-correction-type}, we see that both FineRadScore-GPT-4 and FineRadScore-Opus are able to attain accuracies in the 80s and 90s across correction types on ReFiSco-v1. This trend is mostly true on ReFiSco-v0 with the exception of a few accuracies that are in the high 60s.

\begin{table}[htbp]
    \centering
    \caption{FineRadScore is able to capture the majority of the text that was rewritten and inserted by the radiologist even when findings are shuffled.}
    \label{tab:shuffled-bleu}
    \begin{tabular}{ccccc}
    \toprule
    \bfseries Dataset & \bfseries Model & \bfseries BLEU-1 Scores & \bfseries BLEU-2 Scores & \bfseries BERT Scores \\
    \midrule
    \multirow{ 2}{*}{ReFiSco-v1} & GPT-4 & \bfseries 0.8807 & \bfseries 0.8303 & \bfseries 0.9473 \\
    & Opus & 0.8223 & 0.7634 & 0.9239 \\ 
    \midrule
    \multirow{ 2}{*}{ReFiSco-v0} & GPT-4 & 0.7504 & 0.7030 & 0.8793 \\
    & Opus & \bfseries 0.8074 & \bfseries 0.7533 & \bfseries 0.9071 \\ 
    \bottomrule
    \end{tabular}
\end{table}

Next, we examine whether the text content of the corrections produced by FineRadScore match the text of the corrections produced by the radiologist. Again, we see in Table \ref{tab:shuffled-bleu} that the majority of the text that was rewritten and inserted by the radiologist is captured by FineRadScore even when findings are out of order.

Now, we apply the corrections generated by FineRadScore and see if our corrected candidate report is more aligned with the shuffled ground truth than our original candidate report. As before, we treat the comparison of the uncorrected candidate report and the ground-truth report as our baseline. In Table \ref{tab:shuffled-radcliq}, we see a stark improvement in BLEU scores when applying the corrections generated by FineRadScore to the original candidate report. Due to the improved RadGraph-F1 and RadCliQ scores, FineRadScore's ability to produce high quality corrections appears robust to the shuffling of sentences in the ground truth.

\begin{table}[htbp]
    \centering
    \caption{FineRadScore is able to produce a corrected report that aligns well with the ground-truth report even when findings are shuffled. }
    \label{tab:shuffled-radcliq}
    \begin{tabular}{cccccccc}
    \toprule
    & & \multicolumn{2}{c}{BLEU-2 Scores} & \multicolumn{2}{c}{RadGraph-F1} & \multicolumn{2}{c}{RadCliQ} \\
    \midrule
    \bfseries Dataset & \bfseries Model & \bfseries Base & \bfseries Corrected & \bfseries Base & \bfseries Corrected & \bfseries Base & \bfseries Corrected \\
    \midrule
    \multirow{ 2}{*}{ReFiSco-v1} & GPT-4 & \multirow{ 2}{*}{0.4043} & \bfseries 0.8703 & \multirow{ 2}{*}{0.5337} & \bfseries 0.9363 & \multirow{ 2}{*}{0.0711} & \bfseries -1.162 \\
    & Opus & & 0.8601 & & 0.9238 & & -1.1332 \\ 
    \midrule
    \multirow{ 2}{*}{ReFiSco-v0} & GPT-4 & \multirow{ 2}{*}{0.3824} & 0.8419 & \multirow{ 2}{*}{0.4533} & 0.8610 & \multirow{ 2}{*}{0.2720} & -1.068 \\
    & Opus & & \bfseries 0.8627 & & \bfseries 0.8680 & & \bfseries -1.096 \\ 
    \bottomrule
    \end{tabular}
\end{table}

Finally, we examine FineRadScore's ability to accurately rate the clinical severity of each line in the original report even with a shuffled ground-truth report. Similar as above, the mean absolute errors are all under 1, at around 0.89 for ReFiSco-v1 and 0.67 for ReFiSco-v0.

\subsection{A Paraphrased Candidate Report}
\label{sec:paraphrased-setting}
Now, we want to see how FineRadScore performs when the candidate and ground-truth reports are stylistically distinct. Therefore, we ask GPT-4 to rephrase a subset of report generations in ReFiSco-v1 and had radiologists double-check that the paraphrases maintain the same semantic meaning. We note that in our few-shot prompt, we emphasize to FineRadScore to ignore changes in phrasing and focus on semantic differences that may lead to clinically relevant errors. 

As before, we start by examining if FineRadScore is able to identify the type of correction needed at each line when we paraphrase the candidate report. In Table \ref{tab:paraphrased-correction-type}, we see that FineRadScore struggles more heavily in the ``No Change'' category. For both FineRadScore-Sonnet and FineRadScore-Opus, performance in the ``No Change'' category is quite low. On the other hand, the ``Rewrite'' category has the highest accuracy across models. We conclude that in the paraphrased setting, it seems likely that FineRadScore is prone to over-rewriting. In particular, lines which should not have been rewritten may be getting rewritten due to stylistic differences, rather than differences in clinical meaning.

\begin{table}[htbp]
    \centering
    \caption[FineRadScore Accuracy by Correction Type in Paraphrased Setting.]{FineRadScore falls short in identifying lines that should be left alone. It appears to be prone to over-rewriting. }
    \label{tab:paraphrased-correction-type}
    \begin{tabular}{ccccccc}
    \toprule
    & & \multicolumn{3}{c}{Line Level Correction} & \multicolumn{2}{c}{Report Level Correction} \\
    \midrule
    \bfseries Dataset & \bfseries Model & \bfseries No Change & \bfseries Delete & \bfseries Rewrite & \bfseries Insert & \bfseries No Insert \\
    \midrule
    \multirow{ 2}{*}{ReFiSco-v1} & GPT-4 & \bfseries 72.34\% & 62.96\% & 87.10\% & 78.57\% & 100\% \\
    & Opus & 51.06\% & \bfseries 92.59\% & \bfseries 93.55\% & \bfseries 100.00\% & 100\% \\ 
    \bottomrule
    \end{tabular}
\end{table}

Similarly, when comparing the BLEU scores in Table \ref{tab:paraphrased-bleu}, FineRadScore has lower rewrite and insertion BLEU scores in the paraphrased setting. However, when looking at the lines that FineRadScore rewrites, there are many instances where the rewrites are stylistically different but semantically similar to each other. For example, the radiologist rewrites some line to ``Stable post CABG with intact sternal wires'' and FineRadScore rewrites the same line as ``Sternal wires from coronary artery bypass graft surgery are stable''. As a result, the BLEU score for this rewritten line may be low for FineRadScore, but it is still able to capture the meaning of the correction.

\begin{table}[htbp]
    \centering
    \caption{FineRadScore has lower BLEU scores when comparing the concatenation of rewritten and inserted text to the radiologist.}
    \label{tab:paraphrased-bleu}
    \begin{tabular}{ccccccccc}
    \toprule
    \bfseries Dataset & \bfseries Model & \bfseries BLEU-1 Score & \bfseries BLEU-2 Scores & \bfseries BERT Scores \\
    \midrule
    \multirow{ 2}{*}{ReFiSco-v1} & GPT-4 & \bfseries 0.7060 & \bfseries 0.6640 & \bfseries 0.9014 \\
    & Opus & 0.6224 & 0.5847 & 0.8827 \\ 
    \bottomrule
    \end{tabular}
\end{table}

We see in Table \ref{tab:paraphrased-radcliq} that all models result in corrected candidate reports with better RadGraph-F1 and RadCliQ scores. These results are positive indications that the corrections generated by FineRadScore still produce a high quality corrected report, even in a paraphrased setting. However, they may be prone to producing extraneous corrections based on stylistically differences that are not clinically relevant.

\begin{table}[htbp]
    \centering
    \caption[FineRadScore Corrections Applied RadGraph-F1 and RadCliQ scores in Paraphrased Setting.]{All models result in corrected candidate reports with better RadGraph-F1 and RadCliQ scores in the paraphrased setting.}
    \label{tab:paraphrased-radcliq}
    \begin{tabular}{cccccccc}
    \toprule
    & & \multicolumn{2}{c}{BLEU-2 Scores} & \multicolumn{2}{c}{RadGraph-F1} & \multicolumn{2}{c}{RadCliQ} \\
    \midrule
    \bfseries Dataset & \bfseries Model & \bfseries Base & \bfseries Corrected & \bfseries Base & \bfseries Corrected & \bfseries Base & \bfseries Corrected \\
    \midrule
    \multirow{ 2}{*}{ReFiSco-v1} & GPT-4 & \multirow{ 2}{*}{0.0705} & 0.4753 & \multirow{ 2}{*}{0.2252} & 0.6858 & \multirow{ 2}{*}{0.9811} & -0.4583 \\ 
    & Opus & & \bfseries 0.6150 & & \bfseries 0.7587 & & \bfseries -0.7417 \\ 
    \bottomrule
    \end{tabular}
\end{table}

Finally, we analyze FineRadScore's ability to identify the clinical severity of the error associated with each line (or if no error exists). The mean absolute errors are all under 1 for FineRadScore, at around 0.89 for GPT-4 and 0.58 for Opus. FineRadScore is still able to identify the clinical error severity associated with each line within one category even when the candidate and ground-truth reports are stylistically different.

\section{Alignment with Radiologists on ReXVal}
\label{alignment_radcliq}

RadCliQ (\cite{evaluating-progress-patterns}), G-Rad (\cite{chaves2024training}), and GREEN (\cite{ostmeier2024green}) are some current state-of-the-art methodologies for chest x-ray report evaluation and have been shown to be more aligned with radiologist evaluation than prior metrics. RadCliQ is a composite metric that outputs a single score based off of lexical and clinical metrics for each pair of reports. Given a candidate and ground truth, G-Rad prompts GPT-4 to output the number of errors in each of the six ReXVal error categories found in a candidate report. Additionally, G-Rad instructs GPT-4 to distinguish between clinically significant and clinically insignificant errors, echoing the instructions given to the radiologists while creating the ReXVal dataset. Similarly, given a candidate and ground truth report pair, GREEN outputs a score based on the number of clinically significant errors in each of the six ReXVal error categories. All of these metrics were benchmarked on the ReXVal dataset (\cite{evaluating-progress-patterns}) in order to compare alignment with radiologists with prior evaluation techniques. We therefore also compare FineRadScore's alignment with radiologists using ReXVal. 

However, one key distinction between FineRadScore and prior evaluation metrics is that FineRadScore provides line-level, not report-level, scores for each candidate report. Therefore, we need a way to get a single score for each candidate report from the corrections. Just as in our experiments, when no corrections are found, we set the clinical severity rating to 0. Then, for each of the four severity ratings, we assign the numbers 1 to 4 in order of increasing severity (not actionable, actionable nonurgent error, urgent error, emergent error). We then sum the clinical severity scores of all corrections generated for a given report to get a single FineRadScore score for each report. ReXVal contains radiologist annotations counting the total number of errors in each of six error categories for each report pair. These are the radiologist scores for each report. Using the same bootstrapping technique with 1000 resamples as \cite{evaluating-progress-patterns}, we measure the Kendall tau correlation between FineRadScore's scores and the radiologist scores.

The ReXVal dataset contains radiologist annotations for 200 pairs of candidate and ground-truth reports. Since RadCliQ used part of the ReXVal dataset to train their composite metric, they evaluated on a held out test set of 40 annotated pairs. Therefore, we follow their methodology and compute the Kendall tau correlation on that same held out test set. The Kendall tau correlations, as well as the 95\% confidence intervals are reported in Table \ref{tab:rexval-kendall-tau-radcliq}. BLEU, BERTScore, CheXbert vector similarity, RadGraph, and RadCliQ numbers are taken from \cite{evaluating-progress-patterns}. G-Rad evaluates their methodology on the full ReXVal dataset. Therefore, we use the same technique to obtain Kendall tau scores on the full ReXVal dataset and report the results in Table \ref{tab:rexval-kendall-tau-grad}.

\begin{table}[h!]
  \caption[Kendall tau correlation comparison on ReXVal test set]{FineRadScore has a Kendall tau b correlation comparable to that of RadCliQ and other baseline metrics on ReXVal's held-out test set of 40 data points.}
  \centering
  \begin{tabular}{ll}
  \toprule
   & \bfseries Kendall tau b correlation \\
  \midrule
  BLEU & 0.414 (95\% CI, 0.156 0.635) \\
  BERTScore & 0.505 (95\% CI, 0.273 0.671) \\
  CheXbert & 0.537 (95\% CI, 0.330 0.717) \\
  RadGraph & 0.528 (95\% CI, 0.357 0.687) \\
  RadCliQ & 0.615 (95\% CI, 0.450 0.749) \\
  \midrule
  FineRadScore (GPT-4) & \bfseries 0.701 (95\% CI, 0.523 0.841) \\
  FineRadScore (Claude-3 Opus) & \bfseries 0.737 (95\% CI 0.593 0.850) \\
  \bottomrule
  \end{tabular}
  \label{tab:rexval-kendall-tau-radcliq}
\end{table}

\begin{table}[h!]
\caption[Kendall tau correlation comparison on full ReXVal dataset]{FineRadScore has a Kendall tau b correlation lower, but still with overlapping 95\% CI, compared to that of G-Rad and summed error counts on GREEN on the full ReXVal dataset. FineRadScore has a higher Kendall tau b correlation compared to the GREEN score.}
  \centering
  \begin{tabular}{ll}
  \toprule
   & \bfseries Kendall tau b correlation \\
  \midrule
  G-Rad & 0.76 (95\% CI, 0.70 0.80) \\
  GREEN & 0.63 (95\% CI, 0.56 0.69) \\
  Error counts GREEN & \bfseries 0.79 (95\% CI, 0.74 0.83) \\
  \midrule
  FineRadScore (GPT-4) & 0.700 (95\% CI, 0.631 0.756) \\
  FineRadScore (Claude-3 Opus) & 0.738 (95\% CI, 0.680 0.788) \\
  \bottomrule
  \end{tabular}
  \label{tab:rexval-kendall-tau-grad}
\end{table}

Table \ref{tab:rexval-kendall-tau-radcliq} shows that FineRadScore is just as aligned (if not slightly more aligned) with radiologists as RadCliQ on this held out test set. Table \ref{tab:rexval-kendall-tau-grad} shows that FineRadScore is slightly less aligned as G-Rad on the ReXVal dataset. However, in both of these comparisons, the 95\% CI between FineRadScore and RadCliQ overlap, indicating that FineRadScore is not significantly underperforming on ReXVal compared to existing state of the art evaluation techniques.

The key advantage that FineRadScore brings compared to these existing techniques is its ability to offer much more fine-grained information. Both RadCliQ and G-Rad offer a single score for a candidate report. GREEN also outputs the most representative error explanations for each of the six ReXVal categories. However, FineRadScore goes one step further by giving us the line-by-line corrections, as well as clinical severity ratings. As a result, FineRadScore can help us narrow down which portion of the report is most problematic, as well as what steps can be taken to correct the errors in the candidate report.

\paragraph{Qualitative Analysis of FineRadScore} We experimented with FineRadScore when the candidate and ground-truth reports are semantically identical. When the candidate and ground truths are identical, both FineRadScore-GPT-4 and FineRadScore-Opus output no corrections. When they are shuffled versions of each other, FineRadScore-GPT-4 outputs 0.03 corrections per report and FineRadScore-Opus 0.315 per report. When they are paraphrased versions of each other, FineRadScore-GPT-4 outputs 0.75 corrections per report and FineRadScore-Opus 3.5 per report. The majority of these corrections were labeled as non-actionable. However, these results indicate how FineRadScore is still susceptible to generating corrections purely due to changes in phrasing.

\section{Limitations}
\label{limitations}

Existing automated evaluation metrics such as RadGraph and RadCliQ are currently being used to evaluate FineRadScore. Ideally, we would have a radiologist evaluate the FineRadScore-corrected reports and ensure that they are indeed more aligned with the ground truth report and would recommend future works to do this. However, previously established metrics such as RadCliQ and RadGraph-F1 consistently indicate that the FineRadScore-corrected reports are more aligned with the ground truth. As a result, we believe that this is a strong indication that the corrections are indeed creating reports that are more aligned with the ground truth.

Future work may also focus on improving performance when the ground truth differs significantly stylistically from the candidate report. As shown in the qualitative analysis, FineRadScore is still prone to identifying corrections based off of syntactic differences instead of clinically significant differences. Even so, it is able to achieve comparable performance to state of the art evaluation techniques such as RadCliQ (\cite{evaluating-progress-patterns}) and G-Rad (\cite{chaves2024training}). Therefore, we recommend future work focus on finding a method that can solely identify clinically relevant errors, while ignoring differences in phrasing. We hypothesize that such a finding would improve the performance of LLM-based evaluation techniques across the board.

Additionally, we recommend creating a new dataset using the expert report from the MIMIC-CXR dataset as ground truth. In other words, for a set of model generations, expert annotations regarding the minimum number of corrections needed to transform the model-generated report to be semantically equivalent to the MIMIC-CXR ground-truth report should be collected. In this way, we would be able to evaluate FineRadScore, as well as other evaluation techniques, on this dataset. Evaluation techniques performing strongly on this dataset would likely be most useful to researchers who could now evaluate their own model generations against the MIMIC-CXR ground-truth report.

Finally, we recommend expanding upon the ReFiSco-v1 dataset by collecting duplicate annotations, where multiple radiologists annotate the same report. In doing so, we can use inter-rater disagreement as a baseline, just as we did on ReFiSco-v0 in Section \ref{sec:clinical-severity-ratings}.

\paragraph{Code Availability} In all of our experiments, we used Azure's OpenAI endpoints to access GPT-4. We used Anthropic's endpoints to access Claude-3 Opus. We also signed and adhered to the PhysioNet Data Use Agreement to ensure the confidentiality of the MIMIC-CXR data. Code to run FineRadScore can be found \href{https://github.com/rajpurkarlab/FineRadScore/tree/main}{here}.

\section{Conclusion}
\label{conclusion}

In this work, we introduce FineRadScore, a LLM-based method of obtaining fine-grained evaluation of AI-generated chest X-ray reports. We demonstrate that FineRadScore is able to generate corrections in a way that aligns with radiologists and show how it can be used to evaluate generations in practice. We also demonstrate that FineRadScore's report-level scores align with radiologists' approximately as well as RadCliQ and G-Rad, two state of the art evaluation techniques. However, more work is needed to evaluate pairs of reports that are stylistically dissimilar, in addition to having different content.

% ACKNOWLEDGEMENTS ONLY GO IN THE CAMERA-READY, NOT THE SUBMISSION
\acks{Many thanks to Shirley Shen and Kathy Yu for their help! We also acknowledge support in the form of azure credits from the Microsoft Accelerating Foundation Models Research (AFMR) grant.}

%Do NOT change font size of references or modify the bibliography style
\bibliography{sample}

\newpage
\appendix
\section*{Appendix A.}
\label{AppendixA}

The candidate reports were split into lines using the following regular expression:
\begin{verbatim}
    "(?<!\d)\.(?!\d|$) "
\end{verbatim}

\section{Prompt Instructions}
\label{appendix:prompt-instructions}

We started off each prompt with the same set of prompt instructions, which is a redacted version of the instructions we gave the radiologists. These examples were selected with the goal of covering a wide variety of possible errors in the candidate report, as well as the fact that purely stylistic differences should be ignored. Both few-shot and zero-shot prompts began with these set of instructions. These instructions are shown below.

\begin{verbatim}
Instructions: Return only a json object for this radiology report, with a 
key-value pair for every line. 
Each line starts with a numerical id. 
The key will be the id.
The value will be another JSON object. 

Our goal is to judge prioritizing clinically actionable and urgent findings. 
We are looking to determine the accuracy of report impressions generated 
by models or reports from human radiologists.

For each line, there are two steps:
    1. Correcting the error or inserting the omitted finding: 
        -To make a change to an existing line, copy and paste the text and 
        make your changes. 
        -To add a new line, simply insert a row as necessary. 
        -To entirely delete a line, please enter [delete].
    2. Grading the clinical severity of the error. 
    There are four categories:
    For determining the error urgency, please take into consideration 
    that the patients were seen in an ICU setting.
        - Emergent error: Findings that suggest a need for immediate 
        intervention, with significant impact on patient’s health
        Examples: missed or incorrectly called tension pneumothorax, 
        pneumoperitoneum, significantly malpositioned endotracheal tube
        - Urgent error: Findings where failure to act may adversely affect 
        patient health, that require urgent (but not immediate attention) 
        and that if not acted on, may worsen over time and likely result 
        in an adverse patient outcome
        Examples: missed pneumonia, mildly malpositioned endotracheal tube, 
        incorrect anatomy for pleural effusion
        - Actionable nonurgent error: Findings that likely do not require 
        action in the short term but if not acted upon may reasonably impact 
        the patient’s health.
        Examples: pulmonary nodule which requires follow up, presence of 
        emphysematous changes
        -Not actionable: Findings that reasonably would not likely have an 
        impact on the patient’s
        management. Errors that contain a reference to a nonexistent 
        comparison but otherwise contain the correct information are 
        in this category.
        Examples: chronic appearing rib fracture, describing a 
        consolidation as stable when there is no comparison, 
        description of pulmonary edema as asymmetric rather 
        than bilateral (as this would not change management)
    
Only include semantically relevant errors. For example, if the report 
states "no pleural effusion" and the model generates "no pleural effusion
is seen", this is not an error.
However, incorrect sizes or locations of findings are considered errors. 
Incorrect findings should be removed and omitted findings should be added.

Please label each error with one or more of the 
following 6 error categories:
    1. False prediction of finding
    2. Omission of finding
    3. Incorrect location/position of finding
    4. Incorrect severity of finding
    5. Mention of comparison that is not present in the reference impression
    6. Omission of comparison describing a change from a previous study

Please format your output as a JSON object as shown below.
\end{verbatim}

\subsection{Zero-Shot Prompt}
\label{appendix:zero-shot}

In our zero-shot prompt, we simply provide some rules regarding how we want our output to be formatted, so that we can postprocess the GPT-4 generations. The zero-shot prompt is as follows.

\begin{verbatim}
Results should be a JSON object in the format of: 
{
    "1": {
        "corrections": "Corrected sentence 1 from report.",
        "clinical severity": "severity of error",
        "comments": "any comments regarding the correction",
        "error category": ["list of error categories"]
    },  
    "5": {
        "corrections": "Corrected sentence 5 from report.",
        "clinical severity": "severity of error",
        "comments": "any comments regarding the correction",
        "error category": ["list of error categories"]
    }, 
    "None": {
        "corrections": "Insertion of new sentence for report.",
        "clinical severity": "severity of error",
        "comments": "any comments regarding the correction",
        "error category": ["list of error categories"]
    },
}
\end{verbatim}

\subsection{Few-Shot Prompt} \label{appendix:few-shots}
\begin{figure}[h!]
\includegraphics[width=\textwidth]{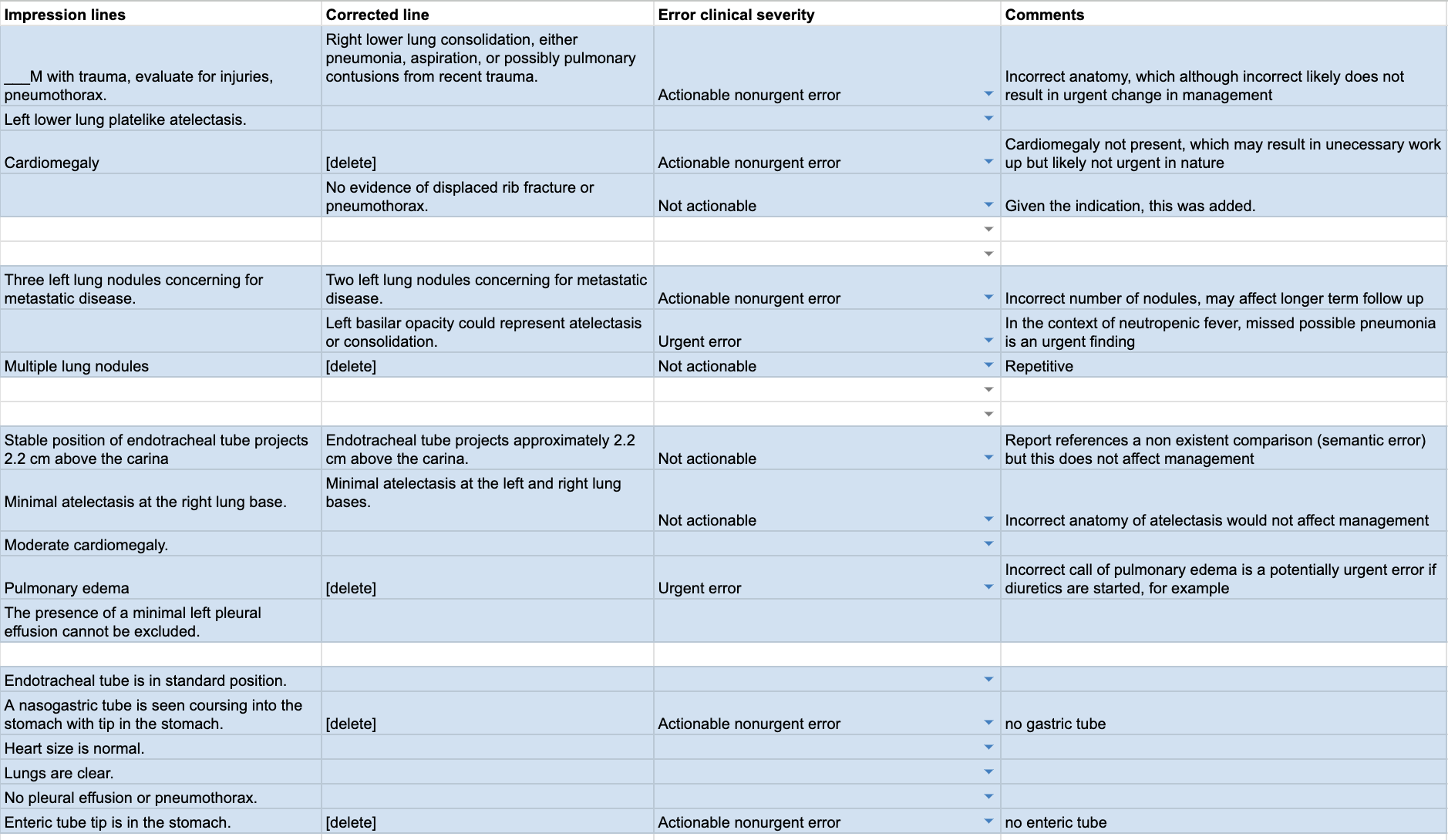}
\centering
\caption[Few-shot Examples Used by FineRadScore]{Few-shot Examples Used by FineRadScore}
\label{fig:few-shot-prompt}
\end{figure}

Below we include the five shots used as part of our few-shot prompt. One of the shots contained no corrections and was designed to emphasize the fact that stylistic changes should be ignored. Three of the examples were taken from the sample examples given to all radiologists at the start of the annotation process. One of the example was taken from one of the ReFiSco-v1 annotations that was excluded from the dataset in all experiments. The examples are shown in Figure \ref{fig:few-shot-prompt}.

The few-shot examples look as follows:
\begin{verbatim}
Example 1:

Generated Text: [0] ___M with trauma, evaluate for injuries, pneumothorax. 
[1] Left lower 
lung platelike atelectasis.
[2] Cardiomegaly.

Ground Truth Text: Right lower lung consolidation, either pneumonia, 
aspiration, or possibly pulmonary contusions from recent trauma. 
No evidence of displaced rib fracture or pneumothorax.

Corrections: 
{
    "0": {
        "corrections": "Right lower lung consolidation, either pneumonia, 
                        aspiration, or 
                        possibly pulmonary contusions from recent trauma.",
        "clinical severity": "Actionable nonurgent error",
        "comments": "Incorrect anatomy, which although incorrect 
                    likely does 
                     not result in urgent change in management",
        "error category": ["Incorrect location/position of finding"]
    },
    "2": {
        "corrections": "[delete]",
        "clinical severity": "Actionable nonurgent error",
        "comments": "Cardiomegaly not present, which may result in 
                    unecessary 
                    work up but likely not urgent in nature",
        "error category": ["False prediction of finding"]
    },
    "None": {
        "corrections": "No evidence of displaced rib fracture 
                        or pneumothorax.",
        "clinical severity": "Not actionable",
        "comments": "Given the indication, this was added.",
        "error category": ["Omission of finding"]
    }
}

Example 2:

Generated Text: 
[0] Three left lung nodules concerning for metastatic disease. 
[1] Multiple lung nodules.

Ground Truth Text: Two left lung nodules concerning for metastatic disease. 
Left basilar opacity could represent atelectasis or consolidation.

Corrections:
{
    "0": {
            "corrections": "Two left lung nodules concerning for 
                            metastatic disease.",
            "clinical severity": "Actionable nonurgent error",
            "comments": "Incorrect number of nodules, may affect longer 
                        term follow up",
            "error category": ["Incorrect location/position of finding", 
                               "Incorrect severity of finding"]
    },
   "None": {
        "corrections": "Left basilar opacity could represent 
                        atelectasis or consolidation.",
        "clinical severity": "Urgent error",
        "comments": "In the context of neutropenic fever, missed possible 
                    pneumonia is an urgent finding",
        "error category": ["Omission of finding"]
   },
   "1": {
        "corrections": "Not actionable",
        "clinical severity": "[delete]",
        "comments": "Repetitive",
        "error category": []
   }
}


Example 3:
Generated Text: 
[0] Stable position of endotracheal tube projects 2.2 cm above the carina. 
[1] Minimal atelectasis at the right lung base. [2] Moderate cardiomegaly. 
[3] Pulmonary edema. 
[4] The presence of a minimal left pleural effusion cannot be excluded.

Ground Truth Text: Endotracheal tube projects approximately 2.2 cm above the 
carina. Minimal atelectasis at the left and right lung bases. Moderate 
cardiomegaly. The presence of a minimal left pleural effusion 
cannot be excluded.

Corrections: 
{
    "0": {
        "corrections": "Endotracheal tube projects approximately 2.2 cm  
                        above the carina.",
        "clinical severity": "Not actionable",
        "comments": "Report references a non existent comparison but this 
                    does not affect management",
        "error category": ["Mention of comparison that is not present in  
        the reference impression"]
    },
    "1": {
        "corrections": "Minimal atelectasis at the left and right lung 
                        bases.",
        "clinical severity": "Not actionable",
        "comments": "Incorrect anatomy of atelectasis would not 
                    affect management",
        "error category": ["Incorrect location/position of finding"]
    },
    "3": {
        "corrections": "[delete]",
        "clinical severity": "Urgent error",
        "comments": "Incorrect call of pulmonary edema is a potentially 
                    urgent error if diuretics are started, 
                    for example",
        "error category": ["False prediction of finding"]
    }
}

Example 4:

Generated Text: [0] Endotracheal tube is in standard position. 
[1] A nasogastric tube is 
seen coursing into the stomach with tip in the stomach. 
[2] Heart size is normal. 
[3] Lungs are clear. 
[4] No pleural effusion or pneumothorax. [5] Enteric tube 
tip is in the stomach.

Ground Truth Text: Endotracheal tube is in standard position. 
Heart size is normal.
Lungs are clear. No pleural effusion or pneumothorax.

Corrections: 
{   "1": {
        "corrections": "[delete]",
        "clinical severity": "Actionable nonurgent error",
        "comments": "no gastric tube",
        "error category": ["False prediction of finding"]
    },
    "5": {
        "corrections": "[delete]",
        "clinical severity": "Actionable nonurgent error",
        "comments": "no enteric tube",
        "error category": ["False prediction of finding"]
    }
}

Example 5:

Generated Text: [0] The lungs are well expanded. [1] There is no pleural 
effusion or pneumothorax. 
[2] The cardiomediastinal and hilar contours are unremarkable.

Ground Truth Text: The lungs are adequately inflated. The contours of the 
cardiomediastinal and hilar regions appear normal. There are no 
indications of pleural effusion or pneumothorax.

Corrections: {}

\end{verbatim}

\section{Clinical Severity Score Graphs}

\begin{figure}
\centering
\includegraphics[width=0.5\textwidth]
{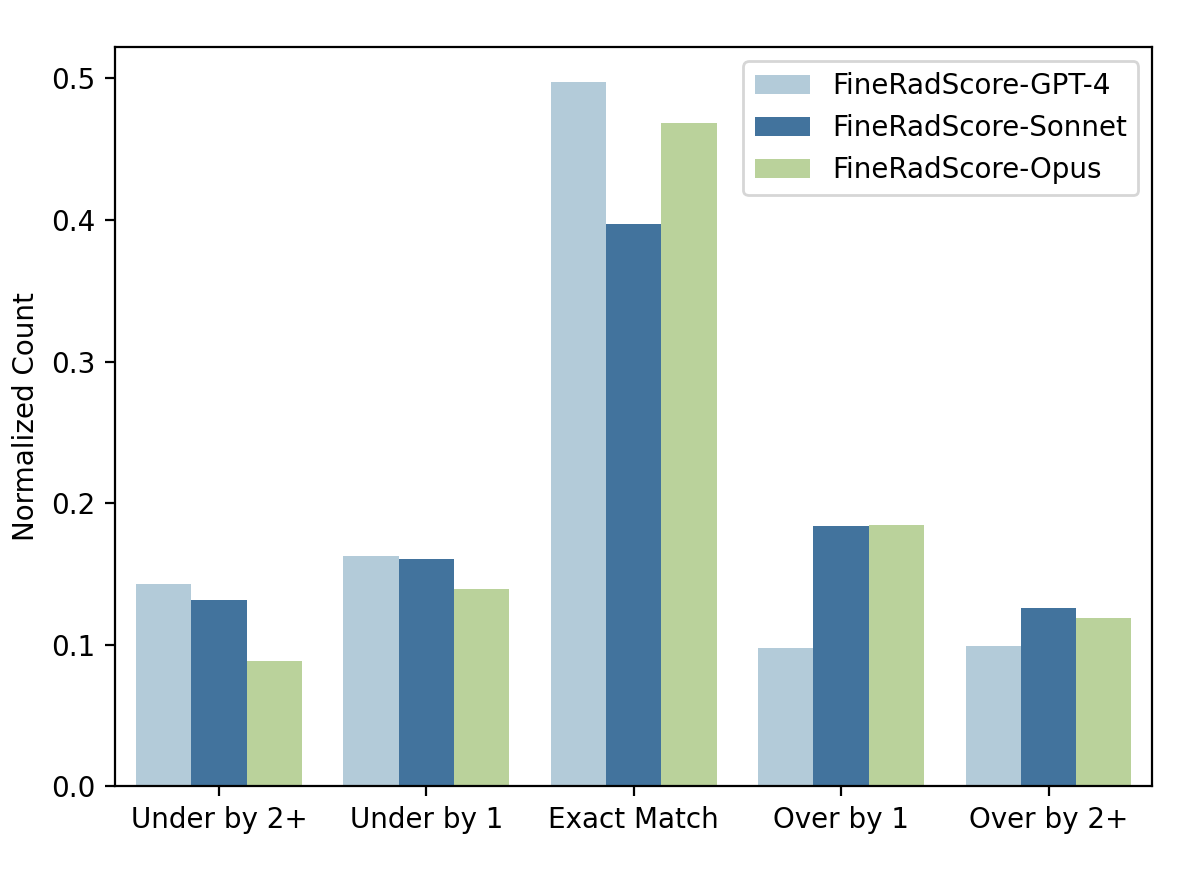}
\caption[FineRadScore Clinical Severity Scores' Accuracies on ReFiSco-v0 with a Shuffled Ground Truth.]{FineRadScore Clinical Severity Scores' Accuracies on ReFiSco-v0 with a Shuffled Ground Truth. FineRadScore's clinical severity scores line up closely with the ground truth radiologist's on ReFiSco-v0. FineRadScore-GPT-4 achieves the best performance in assigning clinical severity scores.}
\label{fig:shuffled-v0-model-comp}
\end{figure}

\begin{figure}
\centering
\includegraphics[width=0.5\textwidth]{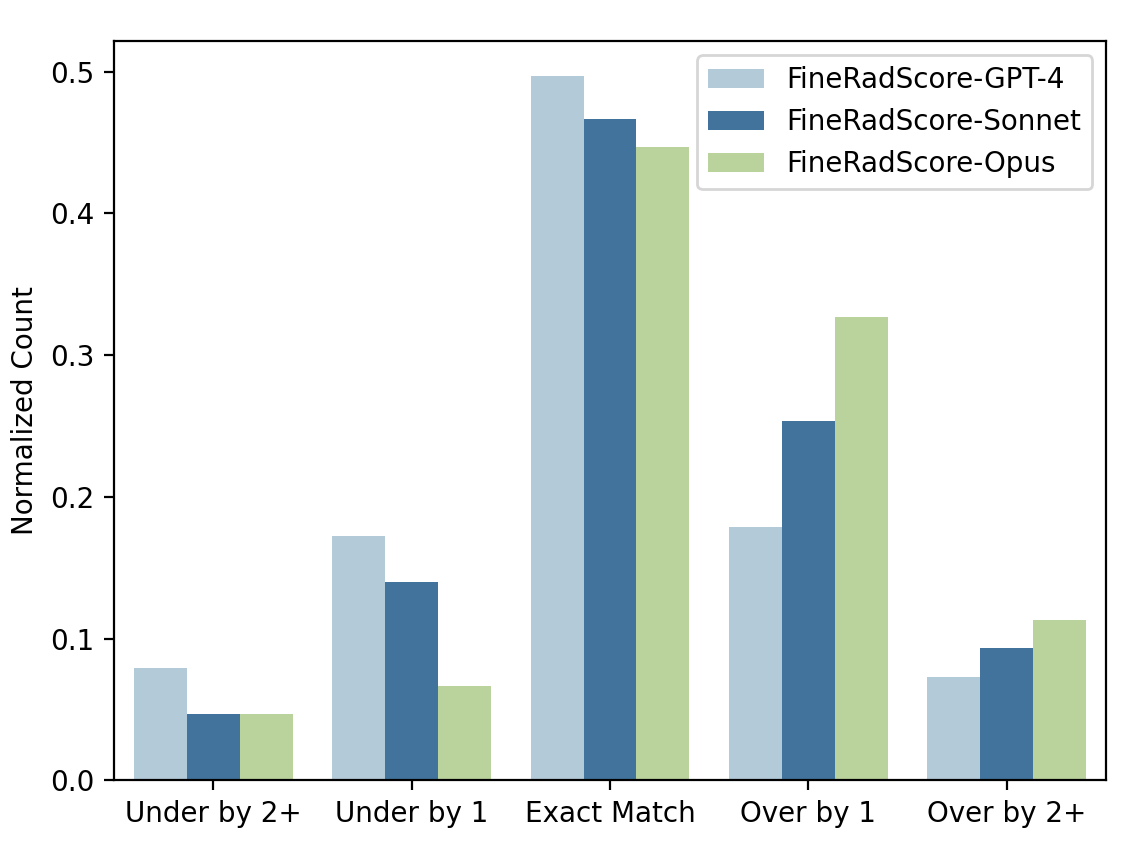}
\caption[FineRadScore Clinical Severity Scores' Accuracies on ReFiSco-v1 with a Paraphrased Ground Truth.]{FineRadScore Clinical Severity Scores' Accuracies on ReFiSco-v1 with a Paraphrased Ground Truth. FineRadScore's clinical severity scores line up closely with the ground truth radiologist's on ReFiSco-v1. FineRadScore-GPT-4 achieves the best performance in assigning clinical severity scores.}
\label{fig:paraphrased-v1-model-comp}
\end{figure}

\newpage

\section{Radiologist Correction Type Graph}

\begin{figure}[h]
\centering
\includegraphics[width=0.5\textwidth]{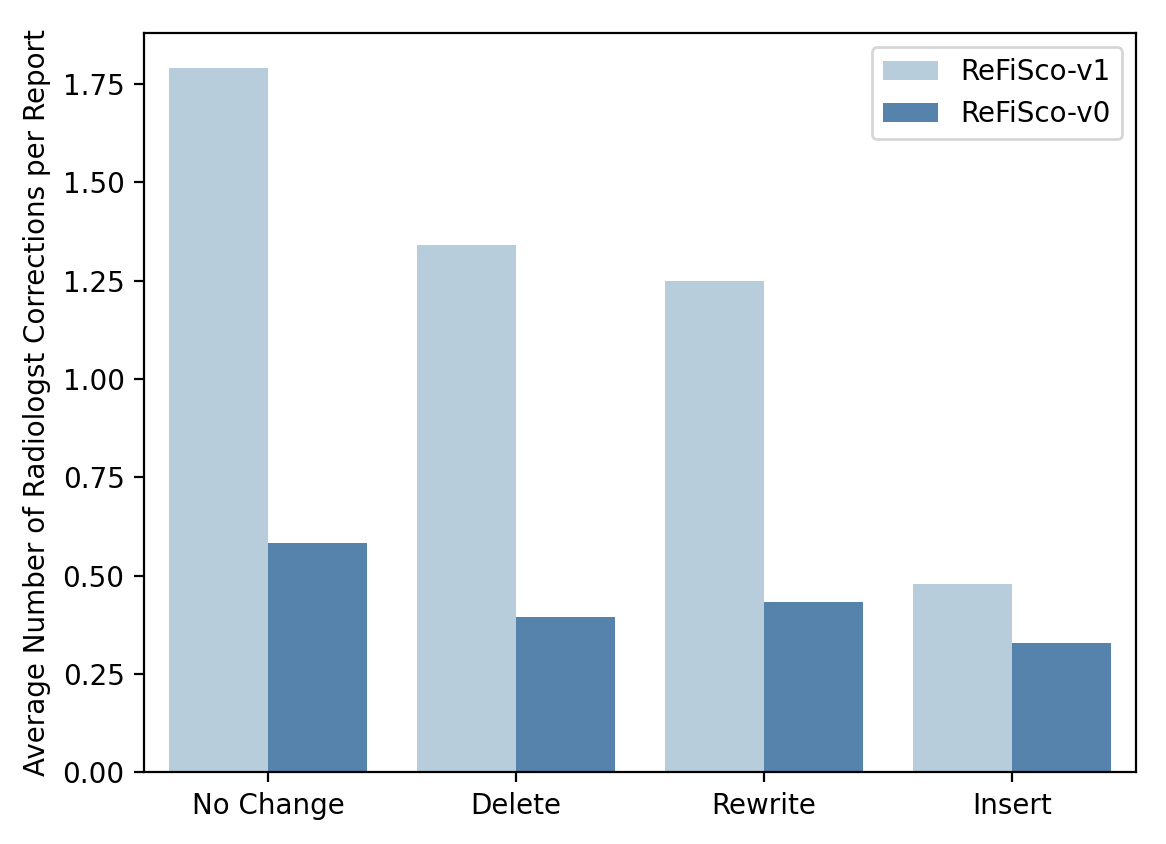}
\caption{Radiologists leave most lines unchanged, and delete or rewrite lines at similar rates. Slightly under half of the reports contain at least one insertion by the radiologist.
\label{fig:5.1_correction_type}}
\end{figure} 

\section{FineRadScore Labeling ReXVal Error Categories}

Overall, FineRadScore still falls short in labeling each correction it generates with the correct ReXVal category. To reiterate, the ReXVal error categories are as follows:
\begin{enumerate}
    \item False prediction of finding
    \item Omission of finding
    \item Incorrect location/position of finding
    \item Incorrect severity of finding
    \item Mention of comparison that is not present in the reference impression
    \item Omission of comparison describing a change from a previous study
\end{enumerate}

For each report in the ReXVal dataset, for each of the categories above, we record whether or not the radiologist found at least one error in that category, as well as whether FineRadScore labeled at least one correction with that category. We use these numbers to compute precision and recall metrics across all six ReXVal categories. As shown in the table below, we see low precision numbers and high recall numbers. These results occur because FineRadScore is over-labeling corrections with these ReXVal error categories. It has a tendency to apply multiple labels to corrections, even when just one fits.

\begin{table}[htbp]
    
    \begin{tabular}{ccccccc}
    \toprule
    & \bfseries Type 1 & \bfseries Type 2 & \bfseries Type 3 & \bfseries Type 4 & \bfseries Type 5 & \bfseries Type 6 \\
    \midrule
    \bfseries Precision & 32.62\% & 64.00\% & 33.33\% & 36.54\% & 40.91\% & 26.39\% \\ 
    \bfseries Recall & 100\% & 100\% & 100\% & 100\% & 100\% & 100\% \\ 
    \bottomrule
    \end{tabular}
\end{table}

\section{FineRadScore Comment Quality}

To evaluate the quality of FineRadScore-GPT-4 generated comments, we ask radiologists to evaluate if a subset of comments were 1) accurate and 2) helpful. The radiologists find 71.98\% of the comments accurate and 79.74\% helpful. We now offer a qualitative breakdown of the types of comments FineRadScore-GPT-4 produces. Some comments simply state that the line that needs to be corrected does not line up with the ground truth, which is technically true, but not informative. Some comments are correct, but referencing the wrong line in the report. Some comments are partially true so still helpful, which led to them getting a ``not accurate'' rating but still helpful.

\end{document}